\documentclass[runningheads]{llncs}

\usepackage[mobile]{eccv}

\usepackage{eccvabbrv}

\usepackage{graphicx}
\usepackage{booktabs}
\usepackage{caption}

\usepackage[accsupp]{axessibility}

\usepackage{hyperref}

\usepackage{orcidlink}

\newcommand{\OURS}{CoMoGen\xspace}
\newcommand{\layer}{Motion Layers\xspace}
\newcommand{\score}{Attention Score}

\begin{document}
\title{CoMoGen: COntrollable MOtion Dynamics \\ and Interactions with Mask-Guided Video GENeration} 
\titlerunning{CoMoGen}

\author{Adil Meric\inst{1} \and Lin Geng Foo\inst{2} \and Mert Kiray\inst{1,3,4} \and Benjamin Busam\inst{1,3,4} \and \\ Rishabh Dabral\inst{2} \and Christian Theobalt\inst{2}}

\authorrunning{Meric et al.}

\institute{Technical University of Munich 
\and Max Planck Institute for Informatics, Saarland Informatics Campus 
\and Munich Center for Machine Learning (MCML)
\and Obsphera}

\maketitle

\begin{center}
  \includegraphics[width=1.\textwidth]{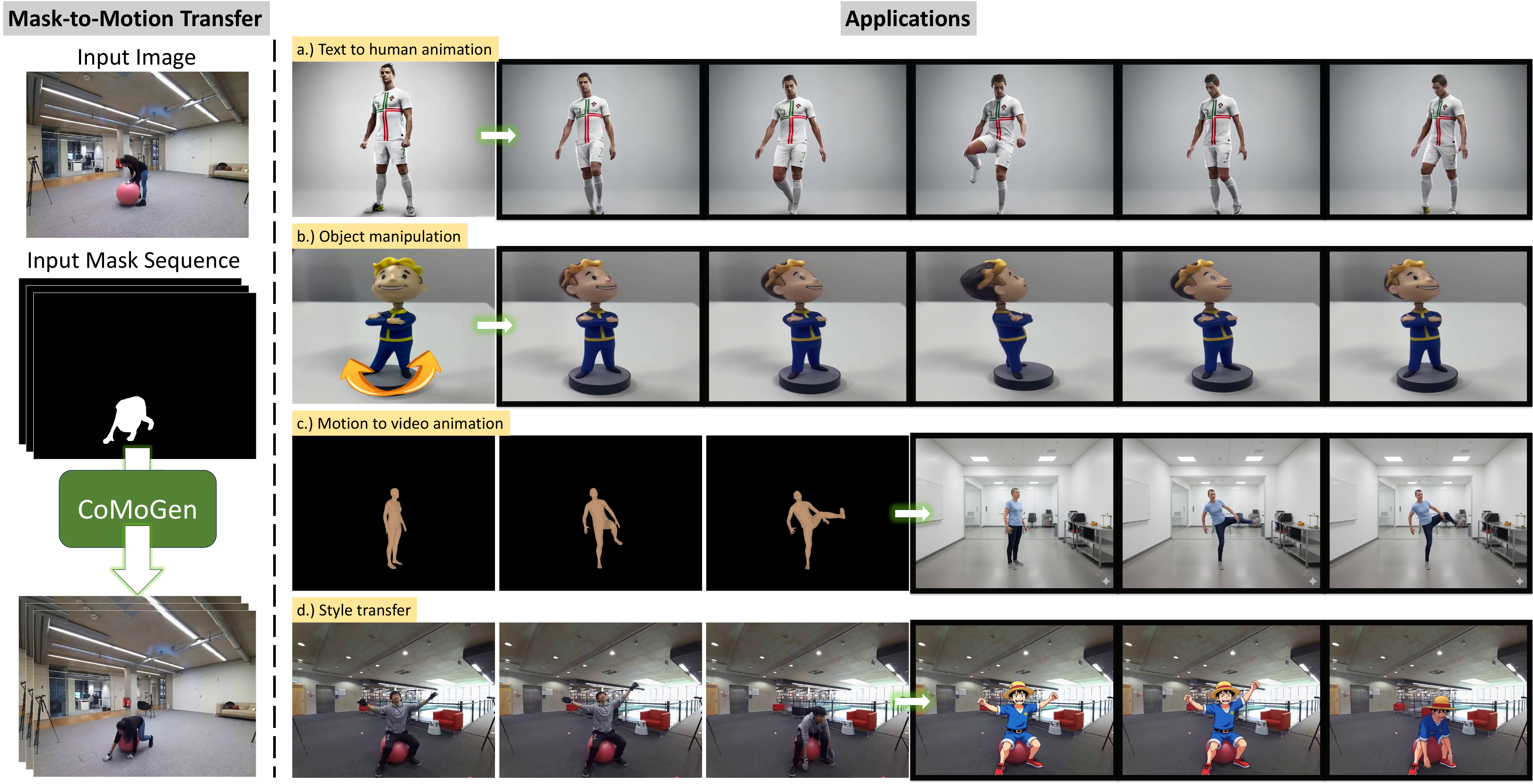}
  \captionof{figure}{\textbf{Overview.} Given a single input image and a binary mask sequence (left), \OURS generates videos where the masked subject follows the steering mask sequence while the model generates plausible interactions with the surroundings. On the right, we show applications: a) Text to human animation from textual motion and rendered masks, b) Object manipulation conditioned on a mask sequence, c) Motion to video animation from generated motion and first frame, and d) Style transfer, animating translated images while preserving motion.}
  \label{fig:teaser}
\end{center}

\begin{abstract}
  We present \OURS, a controllable video generation framework that generates realistic interactive dynamics from a single binary mask sequence conditioned on an input image. \OURS introduces a lightweight MaskAdapter that encodes binary mask sequences into a latent residual signal, injected into the Multi Modal Diffusion Transformer (MMDiT) model through a cosine-weighted schedule. Unlike the hierarchical coarse-to-fine design of UNet architectures, MMDiT operates as a sequence of uniform transformer blocks, making it difficult to identify which layers are responsible for the motion generation. Therefore, we propose a novel way to determine ``Motion Layers'' operating in the attention space of MMDiT. We fine-tune the model by using Low-Rank Adaptation (LoRA) to the \layer, without requiring any architecture change in the MMDiT. This selective adaptation enables our method to focus on motion-critical components, yielding reduced computational cost. Despite its simplicity, \OURS enables precise subject motion and plausible interactions with surrounding humans, objects, and scenes. Comprehensive experiments on different datasets show that \OURS consistently outperforms prior controllable video generation methods and achieves state-of-the-art performance in motion fidelity and perceptual realism. Project page: \href{https://mericadil.github.io/CoMoGen}{mericadil.github.io/CoMoGen}.
  \keywords{Video Generation \and Generative Models \and Conditional Generation}
\end{abstract}

\section{Introduction}
\label{sec:intro}
Recent advances in video generation~\cite{ho2022video, ho2022imagen, svd, videoldm, wan, hunyuanvideo, yang2024cogvideox} have made it possible to generate visually appealing and temporally consistent videos from given text or image inputs. Although these models often exhibit high-quality appearance, making them controllable has emerged as the next frontier for research.
In particular, controlling how objects move and interact remains a central challenge. 
Realistic motion requires an ability to reason about physical relationships between scene entities (humans, objects, and their surroundings) and model sparse video guidance into coherent spatiotemporal dynamics. 
Without this ability, the generated motion may be temporally smooth yet physically implausible, failing to capture contact, collision, or consequential motion in the environment.

Existing controllable video generation frameworks~\cite{zhang2025tora,magicmotion,interdyn,sgi2v,qiu2025freetraj,geng2024motionprompting,shi2024motion,drag,wang2024motionctrl,kuang2024collaborative, yatim2023spacetime,ling2025motionclone,montanaro2024motioncraft,goflow} are mostly built on explicit dense signals such as optical flow or trajectories.
These control signals offer structured guidance but often require extensive pre-processing~\cite{qiu2024freetraj}, domain-specific annotations~\cite{tora, magicmotion}, or additional noise-warping machinery~\cite{goflow}.
Some recent works employ ControlNet~\cite{controlnet} to introduce such control signals~\cite{magicmotion, interdyn, geng2024motionprompting, bahmani2025ac3d, gu2025das}. However, ControlNet typically relies on a trainable copy of the backbone with added zero-initialized layers and produces control residuals across the network, substantially increasing parameter count and training/inference cost, and making controllability dependent on learning how much to inject at each layer.

In this work, we introduce \OURS, a framework that generates videos with controllable motion while exhibiting interactive dynamics from the mask guidance. As shown in Fig.~\ref{fig:teaser} (Mask-to-Motion Transfer), \OURS takes a single input image and a binary mask sequence that specifies where the controlled subject should appear in each frame, and generates a coherent video that follows this motion while having coherent interactions with the scene. 
In other words, beyond following the given mask, the model also produces plausible interactions consistent with the context. We further show several applications (see Fig.~\ref{fig:teaser}, Applications): (a) Text-to-human animation, where human body is reconstructed from single image~\cite{4dhuman} and a motion diffusion model (MDM)~\cite{mdm} generates a mask sequence that drives \OURS; (b) Object manipulation, where hand-driven object motion is captured in a short video, masks are extracted (Grounding DINO~\cite{liu2023grounding} + SAM2~\cite{ravi2024sam2}), and \OURS animates the input image accordingly; (c) Motion to video animation, where an MDM-generated mesh sequence is converted to a realistic first frame via image-to-image transfer~\cite{gemini3} and then animated using the mesh masks; and (d) Style transfer, where we apply image-to-image stylization~\cite{flux1kontext,gemini3} to the input image and generate a stylized video guided by the same masks.

To realize this, we propose a lightweight MaskAdapter that converts a spatiotemporal mask sequence into a latent residual direction. The adapter produces a directional update to the latent variables that biases the generative velocity field toward motions of the moving mask. 
The MaskAdapter is implemented with two convolutional layers followed by a linear projection, and leaves the base model unchanged.
The predicted latent residual direction can be injected into DiT blocks (together with a little LoRA fine-tuning) for mask-guided video generation.

Importantly, to keep our method as lightweight and efficient as possible, we aim to minimize the number of added parameters.
Yet, this is not straightforward; 
in contrast to the hierarchical coarse-to-fine nature of earlier UNet architectures~\cite{svd,videoldm} for diffusion-based video generation, modern Multi Modal Diffusion Transformers (MMDiT)~\cite{flow-ode} perform combined spatial and temporal reasoning throughout a sequence of MMDiT blocks, making it unclear where motion is primarily formed. 
Inspired by insights from DiTCTRL~\cite{ditctrl}, StableFlow~\cite{stableflow}, and TAVID~\cite{kim2025target} we analyze Video-to-Text and Text-to-Video attention alignment between subject tokens and mask regions across inference steps. 
By measuring the alignment between mask regions and attention maps, we identify a sparse subset of DiT blocks and denote them as \layer, which contribute most to dynamic behavior.
We then inject the mask conditioning primarily through these \layer and apply selective Low-Rank Adaptation (LoRA)~\cite{lora} only on them, concentrating learning capacity where motion is formed while keeping the remaining blocks frozen. 
This results in a lower parameter and lower cost control mechanism compared to ControlNet-style designs~\cite{magicmotion,interdyn,geng2024motionprompting,gu2025das} that learn per-layer residual injections across the full network.

Considering that early iterative steps determine global structure and motion while later steps refine details~\cite{attend-excite,cao_2023_masactrl,hertz2023prompttoprompt, meral2024motionflow}, we further introduce a cosine-decayed residual schedule: it weights the mask guided latent shift strongly at early steps and smoothly fades it to zero toward the end. This matches our empirical observation that motion-relevant alignment signals peak in earlier steps and diminish as refinement progresses.

\OURS provides a unified solution for motion control across diverse settings, including human-to-object and object-to-object interactions. The approach is efficient and generalizes across varying spatial resolutions, sequence lengths, and out-of-distribution visual domains. Quantitative and qualitative results on BEHAVE~\cite{behave} and CLEVRER~\cite{clevrer} demonstrate that CoMoGen outperforms prior controllable video generation methods~\cite{interdyn, goflow, magicmotion} and establishes state-of-the-art performance in motion fidelity and generation quality.
Training \OURS only on  videos from the BEHAVE dataset, we demonstrate this capability through diverse applications in Fig.~\ref{fig:teaser}, transferring the same mask-to-motion principle to real-world scenes, enabling object manipulation, and supporting style transfer while preserving motion control. 
Please see the Appendix for the applications' setups and details.

Our main contributions are as follows:
\begin{itemize}
    \item We introduce CoMoGen, a mask-driven controllable video generation framework that enables interactive dynamics from a given input image and a binary mask sequence.
    \item We propose a lightweight MaskAdapter with cosine-weighted latent modulation that enables controllable motion without architectural changes to the base model.
    \item We identify and leverage a subset of transformer blocks, denoted as \layer{}, that are most responsible for motion generation in MMDiT. By injecting conditioning and applying LoRA only on these layers, we reduce computational overhead compared to ControlNet-style full network residual injection while preserving generation quality.
\end{itemize}
\section{Related Work}
\label{sec:related_work}

\paragraph{Video Generation with Diffusion Models.}
Diffusion models have become a dominant paradigm for video synthesis due to their scalability and sample quality. Early works adapted image-based UNets to the video domain~\cite{ho2022video, ho2022imagen}, but suffered from frame drift and limited sequence length. Subsequent approaches introduced temporal attention, latent-space modeling~\cite{svd, guo2023animatediff, videoldm}, and noise alignment~\cite{chang2025warped} to improve temporal consistency; yet UNet approaches often scale poorly due to limited temporal modeling and high spatiotemporal cost. 
To overcome this, recent advances leverage diffusion transformers (DiT)~\cite{yang2024cogvideox, wan, hunyuanvideo, opensora}, enabling longer videos, higher resolutions, and better motions. Latest models match the power of DiT with flow-matching and generate impressive results~\cite{wan, hunyuanvideo}. 
However, these generative video models are typically conditioned only on text or images with limited control over the generated motion and interactions.
Recently, DiTCtrl~\cite{ditctrl} was proposed to address limitations of short video generation by performing MMDiT attention analysis to improve consistency in a tuning-free manner. TAVID~\cite{kim2025target} introduces a loss in attention space for target-to-mask alignment. Stableflow~\cite{stableflow} determines vital layers, layers that are essential for image formation, by measuring the deviation in image content when bypassing each layer individually. Inspired by DiTCtrl, TAVID, and StableFlow in the image domain, we analyze attention layers to locate \layer, and then combine LoRA with a MaskAdapter to achieve efficient, lightweight controllability.

\paragraph{Controllable Video Generation via Explicit Signals.}
To address controllability, several methods incorporate external motion signals. Trajectory-based approaches encode object paths as sparse points, bounding boxes, or dense masks and integrate them into video generation pipelines~\cite{tora, magicmotion, geng2025motion}. Control signals are fused through dedicated modules like trajectory encoders~\cite{tora}, ControlNet-style adapters~\cite{magicmotion, interdyn}, or attention-guided layers~\cite{wang2023videocomposer, yin2023dragnuwa}. 
In particular, several works \cite{interdyn,alhaija2025cosmos} focus on mask-conditioned video generation via ControlNet adapters, yet ControlNet adapters contain large numbers of parameters which often require considerably large amounts of training data.
Others explore flow-guided control by warping the noise in temporal dimension~\cite{goflow}, enforcing temporal correlation while maintaining flexibility. FreeTraj~\cite{qiu2024freetraj} achieves trajectory control without tuning by adapting the sampling schedule. 
While effective, these approaches often rely on curated annotations or handcrafted motion formats. Moreover, the fine-tuning approaches often require a large amount of carefully curated data together with many trainable parameters, while the training-free methods face issues handling interactions of objects with the surroundings.
Differently, we aim to perform mask-guided video generation with dynamics and interactions, in a manner that is efficient, architecture independent, and generalizable.

\paragraph{Interactive Dynamics and Physical Modeling.}
Modeling object interactions and physical causality poses additional challenges. InterDyn~\cite{interdyn} introduces a binary mask sequence as a minimal control signal to drive physical interactions, enabling realistic modeling of physical events such as collisions and contact propagation by training a ControlNet style condition network. CosHand~\cite{coshand} takes a more limited approach focused on static hand-object transitions. 
However, these methods often require large amounts of curated data and add many additional parameters to their model for fine-tuning, which we eschew in our work.
Recent work also explores video generation with the aid of physical simulators: MotionCraft~\cite{montanaro2024motioncraft} combines a physical simulator with an image generation model for video generation, and MotionCtrl~\cite{wang2024motionctrl} blends multiple motion controllers for better generalization while using training data from a physical simulator.
Yet, these methods using physical simulation engines are often limited in capability, e.g., facing difficulty in  modeling human motions, object deformations, etc. 
In contrast,  \OURS  captures interactive effects without depending on simulation engines, which has the advantage of simplicity and does not face limitations brought by the physical simulators.

\begin{figure*}[ht]
  \centering
  \includegraphics[width=0.8\textwidth]{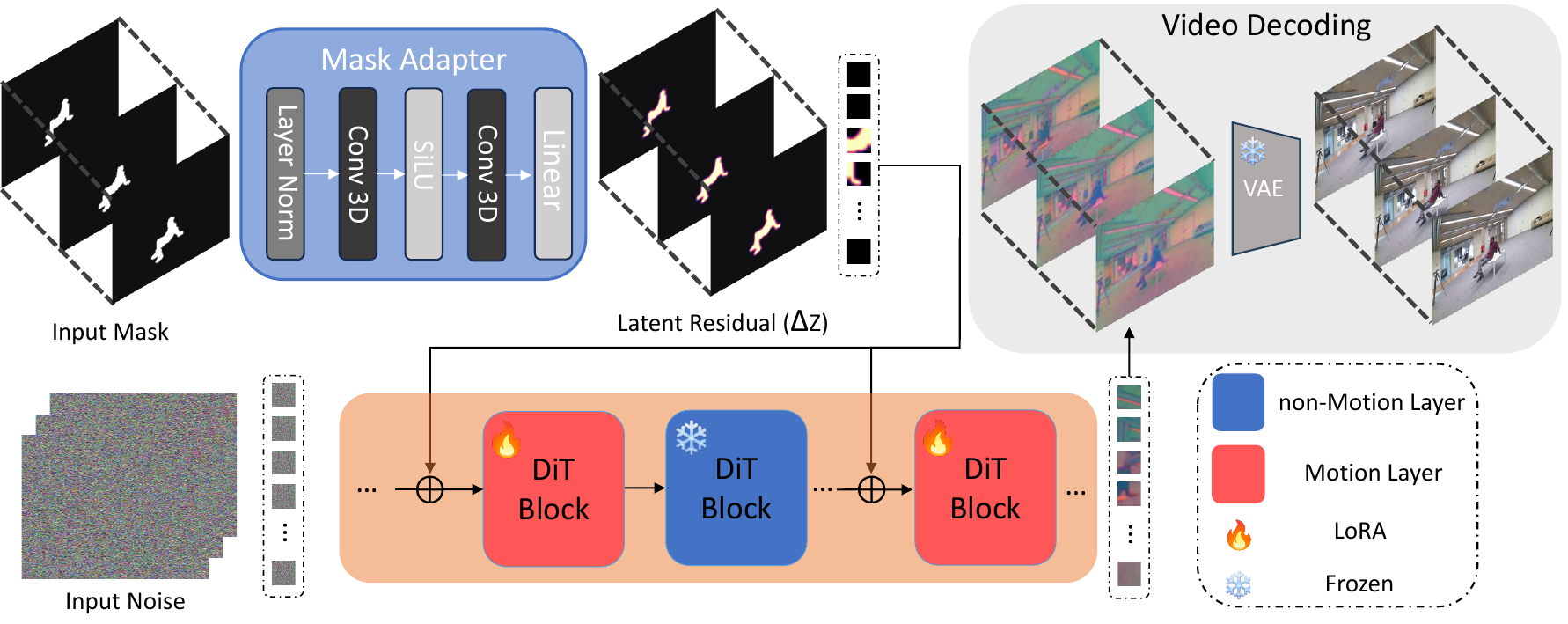}
  \caption{
  \textbf{Framework overview.} Given a spatiotemporal binary mask sequence, we first downsample it to latent resolution and project it into a latent residual $\Delta Z$ using a lightweight MaskAdapter. During image-to-video generation, we inject this residual by modulating the current noised video latent at the transformer input, and propagate the conditioning through only the identified \layer{} (DiT Blocks marked in red), while keeping the remaining blocks frozen (DiT Blocks marked in blue). LoRA modules are applied exclusively to the \layer{} to adapt motion behavior efficiently without disturbing the backbone’s spatial priors. The strength of the injected residual is controlled by a reverse cosine schedule, enforcing strong mask guidance in early denoising steps, where coarse spatial structure and motion are formed, and gradually fading to zero toward the final steps to preserve visual quality. 
  }
  \label{fig:framework}
\end{figure*}

\section{Preliminaries}

\label{sec:prelim}

\subsection{Diffusion Transformers (DiT) and Multi Modal DiT}
Classical diffusion models~\cite{ddpm, ddim, ldm, imagen, svd} rely on UNet backbones with encoder-decoder structures and skip connections, and incorporate conditioning via cross-attention~\cite{imagen, ldm, svd}. While effective for images, these designs often scale poorly to videos due to limited temporal modeling and high spatiotemporal cost.

Diffusion Transformers (DiT)~\cite{DiT} replace convolutional UNets with transformer blocks operating on latent tokens, offering global receptive fields and strong text-image alignment; however, the contribution of individual layers becomes less explicit. Building on this, Multi Modal Diffusion Transformers (MMDiT) \cite{flow-ode,hunyuanvideo} jointly processes multimodal tokens (e.g., text and video latents) by combining self-attention and cross-attention. This formulation is adapted in many modern video generators~\cite{hunyuanvideo, yang2024cogvideox, wan}.

\subsection{Rectified Flow Matching}

Rectified flow~\cite{flow-ode} is an iterative denoising technique, which learns a velocity field $v_\theta(x_t, t)$, often parameterized by a deep model $\theta$. The velocity field maps a noisy latent $x_t = (1-t)x_0 + t\epsilon$ toward clean data $x_0$, where $t$ is the time step and $\epsilon$ is random noise.  
Formally, the training loss is:
\begin{equation}
\mathcal{L}_{\text{FM}} =
\mathbb{E}_{x_0, \epsilon, t}
\bigl[\|v_\theta(x_t, t) - (x_0 - \epsilon)\|_2^2\bigr].
\end{equation}
Unlike diffusion, flow matching follows straight trajectories between data and noise, causing faster convergence and stable gradients which is used in state-of-the-art video generation models~\cite{wan,hunyuanvideo}.
\begin{figure*}[t]
  \centering
  \begin{minipage}[t]{0.49\textwidth}
    \centering
    \includegraphics[width=\linewidth]{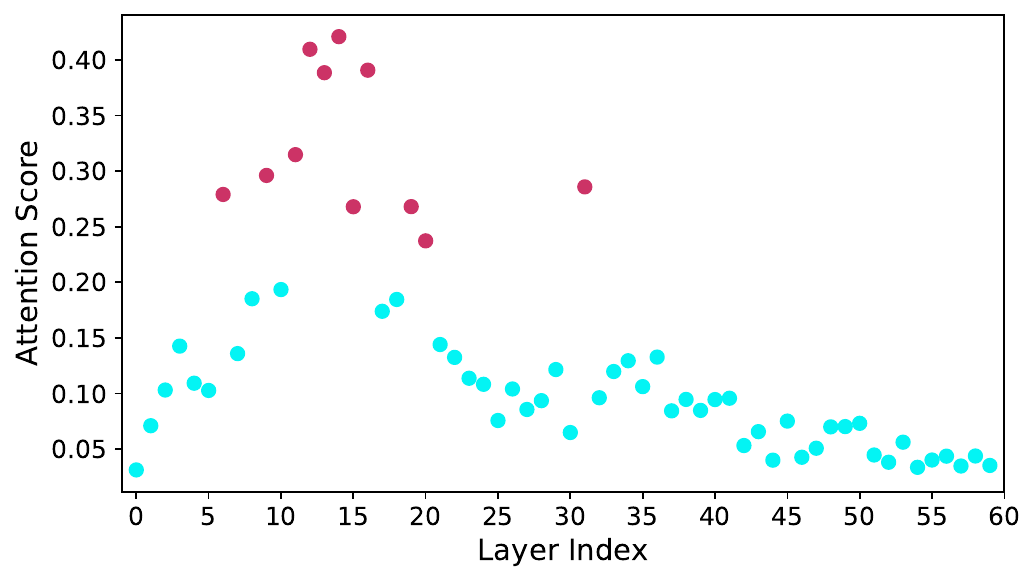}
    \vspace{-3mm}
    \caption{\textbf{Attention score over layers.} \layer{} (marked in red) show consistently higher subject-mask alignment than Non-\layer{} (marked in blue).}
    \label{fig:layer_score}
  \end{minipage}\hfill
  \begin{minipage}[t]{0.48\textwidth}
    \centering
    \includegraphics[width=\linewidth]{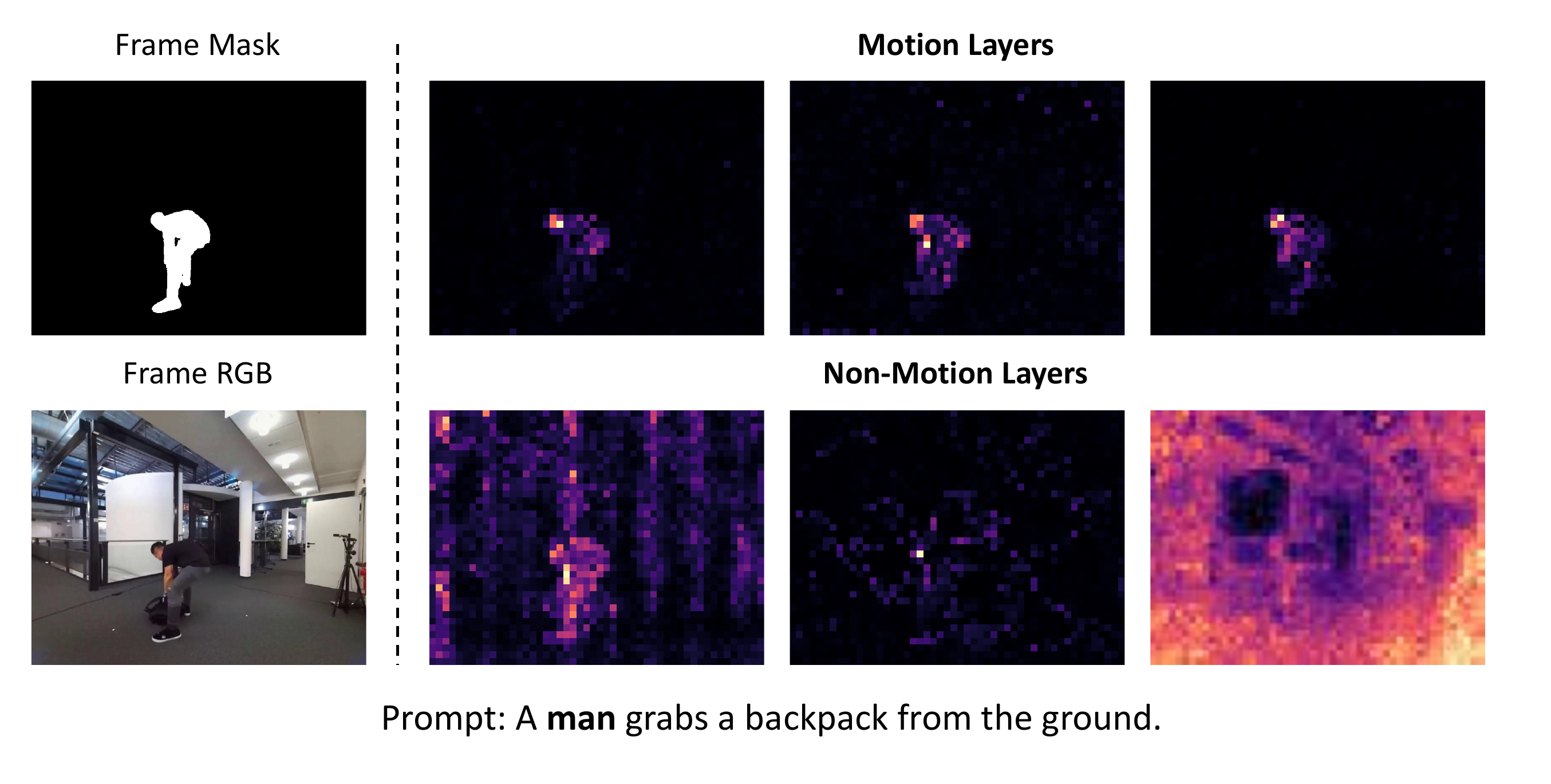}
    \vspace{-3mm}
    \caption{\textbf{Attention visualization.} \layer{} concentrate attention on the masked subject token; non-motion layers show weak alignment.}
    \label{fig:attn_vis}
  \end{minipage}
  \vspace{-2mm}
\end{figure*}

\section{Method}
We address interactive dynamics between objects given a controllable mask signal. Given an input image $I \in \mathbb{R}^{H \times W \times 3}$ and a sequence of binary masks $M \in \mathbb{R}^{T \times H \times W}$, \OURS generates a video $V \in \mathbb{R}^{T \times H \times W \times 3}$, in which the target subject (a person) follow the motion defined by the provided mask sequence, resulting in coherent and physically plausible interactions with the surrounding environment. 
\par
To this end, we employ a MaskAdapter that maps the input binary masks to a latent residual (delta), representing the direction of change in the latent space (see Fig.~\ref{fig:framework}).
However, instead of learning the MaskAdapter residuals for all layers, we first identify the \textit{vital} layers responsible for mask-video correlation, dubbed \layer.
Building on our analysis of \layer, we apply LoRA~\cite{lora} only to these transformer blocks enabling efficient fine-tuning focused on motion-critical components. This selective adaptation preserves the base model’s strong spatial priors while enhancing controllable motion generation.
%
We further introduce a reverse cosine scheduler to modulate the strength of this delta across the steps of the flow matching process: its weight starts at $1$ for the first step and gradually decays to $0$ by the final step. 
\par
We first describe how we identify the \layer important for video motion generation in Section~\ref{sec:layer_importance}. We then present the proposed MaskAdapter architecture in Section~\ref{sec:arch}, including our layer-selective residual injection strategy and the corresponding LoRA training restricted to these \layer. Finally, we detail our reverse cosine scheduler for cosine-weighted delta injection which is designed to preserve the base model’s fine detail capacity while enabling controllable motion in Section~\ref{sec:cos}.

\subsection{Identifying \layer}
\label{sec:layer_importance}

Building on the MMDiT architecture, we first examine how motion is internally represented within the transformer layers. We hypothesize that while each DiT block contributes to both spatial and temporal reasoning, their relative importance varies. To efficiently control motion without injecting to the full model, we identify a subset of layers, denoted as \layer, that are most the critical for motion dynamics. 
Given an input image $I$, subject token $s$ in the text condition, and a binary mask of the subject region $M_f$, we obtain the initial latent noise and track the attention maps $A^t_{\ell,s,f}$ between the subject token and spatial latent features throughout generation, where $f$ is the frame number, $t$ is the timestep and $\ell$ denotes the layer number. 
Following the findings of~\cite{ditctrl}, we consider both the text-to-video and video-to-text attention directions, and compute their average to capture bidirectional dependency between visual and textual modalities. 
We then compute similarity score, $S_\ell$, between the binary mask $M_f$ and the attention map for each layer. 
This is computed by averaging $A^t_{\ell,s,f}$ through all attention heads and applying frame-wise normalization. 
Finally, we sum the normalized values inside and outside the mask to calculate \score{} as follows:
\begin{equation}
S_\ell = \frac{\sum_{f, t} M_f \otimes A^t_{\ell,s,f}}{\sum_{f, t} A^t_{\ell,s,f}},
\end{equation}

Intuitively, layers with higher $S_\ell$ exhibit stronger focus on the entirety of the moving subjects across all frames, strongly attending to their full motions (video-to-text attention) and also driving them (text-to-video attention), indicating their greater contribution towards the generated motions.
We show the individual layer analysis in Fig. \ref{fig:layer_score} and mark the \layer as red. We also visualize the attention maps of \layer and selected Non-\layer in Fig.~\ref{fig:attn_vis}.
As shown in Fig.~\ref{fig:layer_score}, we observe a clear separation in attention score after the $11^{\text{th}}$ layer, and we therefore set the top-11 layers as \layer. See the supplementary material for the box plot with error bars.

\begin{figure}[t]
  \centering
  \includegraphics[width=.9\columnwidth]{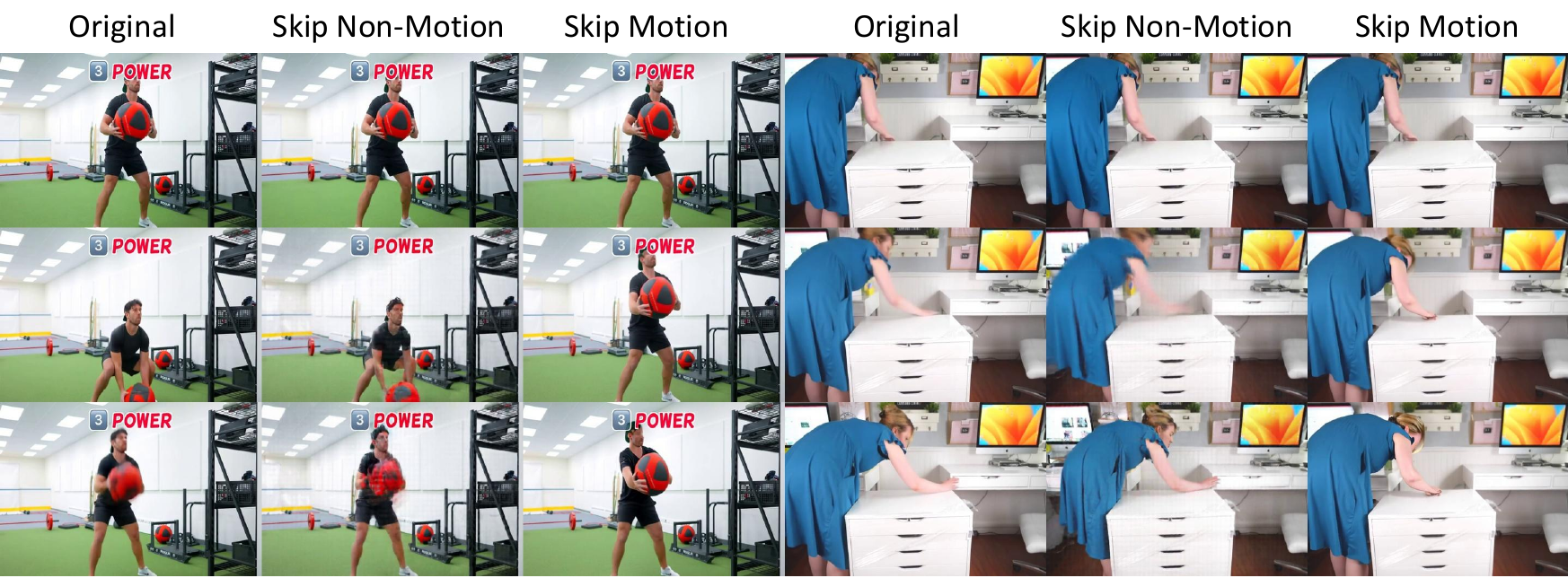}
  \caption{\textbf{Layer skipping.} Skipping Non-\layer{} mainly introduces artifacts, whereas skipping \layer{} disrupts motion dynamics and temporal coherence.}
  \label{fig:layer_skip}
\end{figure}

To experimentally validate that the identified \layer are responsible for motion formation, we conduct a controlled layer-skipping experiment using online videos of InterPose~\cite{interpose} dataset. We first generate 60 videos using our full model. For each video, we then sample two sets of three layers: three randomly selected layers from \layer, and three randomly selected layers from 11 lowest scoring layers among Non-\layer. Using the same initial random noise, we generate two additional outputs by skipping the selected layers in each set, resulting in a triplet per sample: the original generated video, the \layer skipped video, and the Non-\layer skipped video. This design allows us to isolate the effect of removing motion-relevant layers and removing layers that are less relevant according to the Attention Score.

For evaluation, we extract the subject mask from the generated videos using Grounding DINO~\cite{liu2023grounding} and Segment Anything 2~\cite{ravi2024sam2}, and consider the original video masks as pseudo-ground truth tracking labels.
We then evaluate the generated videos with tracking metrics that measure \textit{motion consistency}, reporting $\mathcal{J}$, $\mathcal{F}$~\cite{Perazzi2016}, and HOTA~\cite{luiten2020trackeval,luiten2020IJCV} scores. See the supplementary material for more details about the metrics.
As shown in Table~\ref{tab:JF_score}, skipping Non-\layer results in significantly higher tracking scores than skipping \layer, indicating that removing Non-\layer has less effect on motion formulation, whereas removing \layer disrupts the structure of the subject motion.

In addition to motion consistency, we evaluate \textit{semantic alignment} between the input text description and the generated videos using the VQA score~\cite{lin2024evaluating,Qwen2.5-VL}. Table~\ref{tab:vqa_score} shows that the original generations achieve the highest alignment, and that skipping Non-\layer produces only a small difference as compared to the original results. 
In contrast, skipping \layer leads to a significant drop in VQA score, suggesting that these layers are also important for text-to-motion alignment.

Finally, in Fig.~\ref{fig:layer_skip} we observe that, while skipping Non-\layer introduces visual degradation, such as noise or local patch artifacts, the overall motion remains mostly consistent with the original generation. In contrast, skipping \layer changes the motion pattern drastically and can lead to critically degraded outputs, including incorrect dynamics or unrealistic results. Together, these results provide direct evidence that the proposed \layer identification captures layers that are critical for video motion formation.

Motivated by this analysis, we apply the mask-condition injection to \layer, focusing learning capacity on motion-related blocks while preserving the spatial appearance priors of the base model.

\begin{table*}[t]
  \centering
  \begin{minipage}[t]{0.49\textwidth}
    \centering
    \scalebox{0.8}{
      \begin{tabular}{l cc cc c c}
\hline
Method & $\mathcal{J}\uparrow$& $\mathcal{F}\uparrow$ & $\mathcal{J}\&\mathcal{F}\uparrow$ & HOTA$\uparrow$\\
\midrule
Skip Motion Layers     & 60.1 & 59.9 & 60.0 & 57.3\\
Skip Non-Motion Layers & \textbf{77.7} & \textbf{81.1} & \textbf{79.4} & \textbf{82.3}\\
\bottomrule
\end{tabular}%
    }
    \vspace{2mm}
    \caption{\textbf{Layer analysis for motion formation.} We generate videos skipping the randomly selected 3 \layer{} and Non-\layer{}, and report the tracking metrics.}
    \label{tab:JF_score}
  \end{minipage}\hfill
  \begin{minipage}[t]{0.49\textwidth}
    \centering
    \scalebox{0.9}{
      \begin{tabular}{l c}
\hline
\textbf{Video Type} & \textbf{VQA Score} \\
\hline
Generated Video & \textbf{0.4540} \\
Skip Motion Layers            & 0.2742 \\
Skip Non-Motion Layers        & 0.4068 \\
\hline
\end{tabular}
    }
    \vspace{2mm}
    \caption{\textbf{Layer analysis on text-to-video metric.} We generate video triplets, using the base model, skipping the  selected 3 \layer{}, Non-\layer{}, and report text-to-video alignment.}
    \label{tab:vqa_score}
  \end{minipage}
\end{table*}

\subsection{Mask Adapter}
\label{sec:arch}
The Mask Adapter $\mathcal{MA}(\cdot)$ learns to project binary spatiotemporal masks into latent-space delta features that condition 
the video generation process. 
Given a binary mask tensor $M \in \{0,1\}^{B \times 1 \times T \times H \times W}$, we first map it to the latent dimension $Z_t \in \mathbb{R}^{B \times C \times \frac{T}{4} \times \frac{H}{8} \times \frac{W}{8}}$. We spatially downsample the mask and temporally compress it by applying a logical OR operation across every non-overlapping window of four consecutive frames.
Then we input the downscaled and mean-normalized mask through the Mask Adapter $\mathcal{MA}(\cdot)$, 
which predicts the latent-space residual $\Delta Z = \mathcal{MA}(M)$ that modulates the dynamics of the video generation process. 
We visualize the mask projections in Fig. \ref{fig:framework}, demonstrating that it learns to affect the subject's occupied region, and successfully adapts the surrounding interactions.

\noindent \textbf{LoRA on \layer:} To adapt motion behavior while minimizing overfitting, we apply LoRA into each attention projection 
across \layer. Only LoRA and the MaskAdapter parameters are updated during training, while the base model remains frozen. 

\subsection{Cosine-weighted Latent Injection} 
\label{sec:cos}
While implementing LoRA on motion layers is already effective, we observe that the model is also sensitive to when the condition is injected in the denoising process.
Intuitively, we hypothesize that earlier denoising steps (close to Gaussian noise) are more crucial for mask-motion alignment, as they determine the large motions (i.e., rough movements shapes and silhouettes) that the subsequent denoising steps refine.
Let the step index $s \in \{0, \dots, \tau - 1\}$ and normalized timestep $t_s = \frac{s}{\tau-1}$. 
The cosine scheduler weight is defined as
\begin{equation}
w_s = \frac{1}{2}\bigl(1 + \cos(\pi t_s)\bigr),
\end{equation}
so that $w_0 = 1$ and $w_{\tau-1} = 0$. 
At step $s$, the latent update is
\begin{equation}
\widetilde{Z}_t^{(s)} = Z_t^{(s)} + w_s \cdot \Delta Z.
\end{equation}
This schedule ensures that mask-conditioned motion guidance is strong in early steps and smoothly fades as the velocity prediction converges.

\section{Experiments}
\label{sec:experiments}

We evaluate \OURS under mask guidance and ask whether a single binary mask sequence of a moving subject is sufficient to induce realistic interactions with the surrounding scene. Our evaluation covers two complementary settings that probe controllability, interaction quality, and overall video realism. In Section~\ref{sec:clevrer}, we analyze \textbf{object interactions} in a synthetic environment using CLEVRER~\cite{clevrer}. In Section~\ref{sec:behave}, we study \textbf{human-object interactions} and measure how well the model reproduces subject motion and physical contact between humans and manipulated objects. We compare our method with  state-of-the-art controllable methods: InterDyn~\cite{interdyn}, Go-with-the-flow (GoFlow)~\cite{goflow}, and MagicMotion~\cite{magicmotion}. 
We report results of InterDyn*, where we reproduce their method for both datasets.

\subsection{Dataset}
We use two different datasets CLEVRER~\cite{clevrer} and BEHAVE~\cite{behave}, and train two different models on each dataset.
CLEVRER~\cite{clevrer} provides synthetic videos with multiple objects undergoing collision events. CLEVRER provides 10k training and 5k validation videos. 
We use all videos in the training dataset and use a subset of 50 videos of the validation dataset. 
We use object-centric CLEVRER dataset to test the force propagation, counter factual dynamics and motion fidelity similar to InterDyn~\cite{interdyn}.
BEHAVE~\cite{behave} contains human-object interactions captured in real scenes, including strong occlusions, non-rigid human motion, and contact rich manipulation. We use 4k video subset of the dataset for training and 50 video subset for validation. In our setting, we provide only the human mask as a control and interaction signal, and we evaluate whether the generated videos preserve temporally coherent human motion while producing plausible responses.

\begin{figure*}[t]
  \centering
  \includegraphics[width=\textwidth]{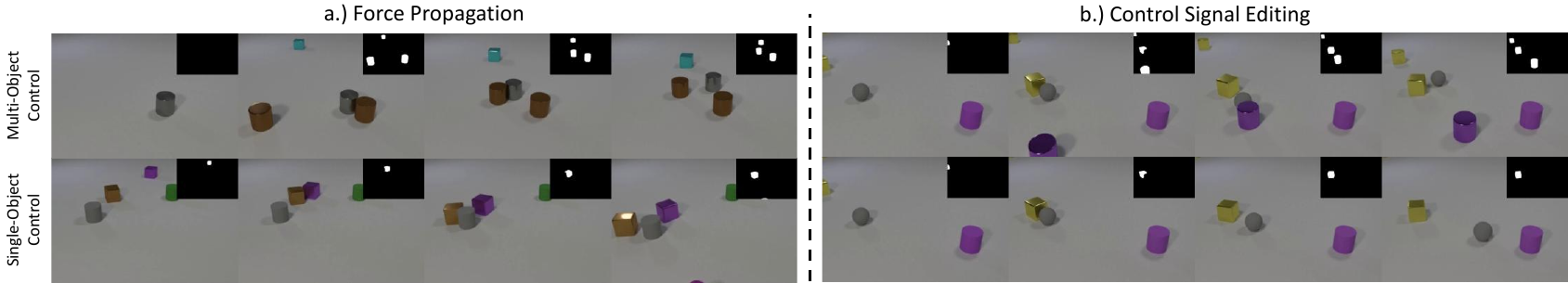}
  \caption{\textbf{Results on human--object interaction.} We present generation results for (a) Force Propagation and (b) Control Signal Editing. For (b), we use the same input image with different control signals.}
  \label{fig:results_clevrer}
\end{figure*}

\begin{figure*}[t]
  \centering
  \includegraphics[width=.85\textwidth]{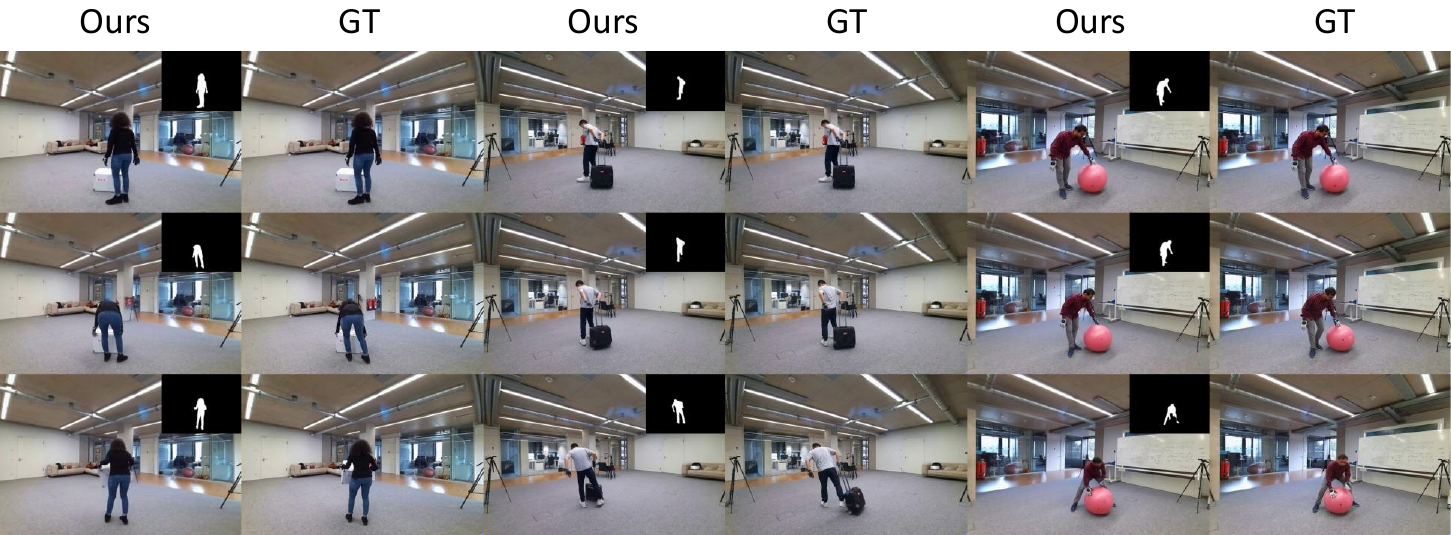}
  \caption{\textbf{Results on human-object interaction.} We present generation results and the corresponding ground truths for different objects (box, luggage bag, and ball).
  }
  \label{fig:results_behave}
\end{figure*}

\subsection{Object Collisions and Reaction Dynamics}
\label{sec:clevrer}

\begin{figure*}[t]
    \centering
        \includegraphics[width=\textwidth]{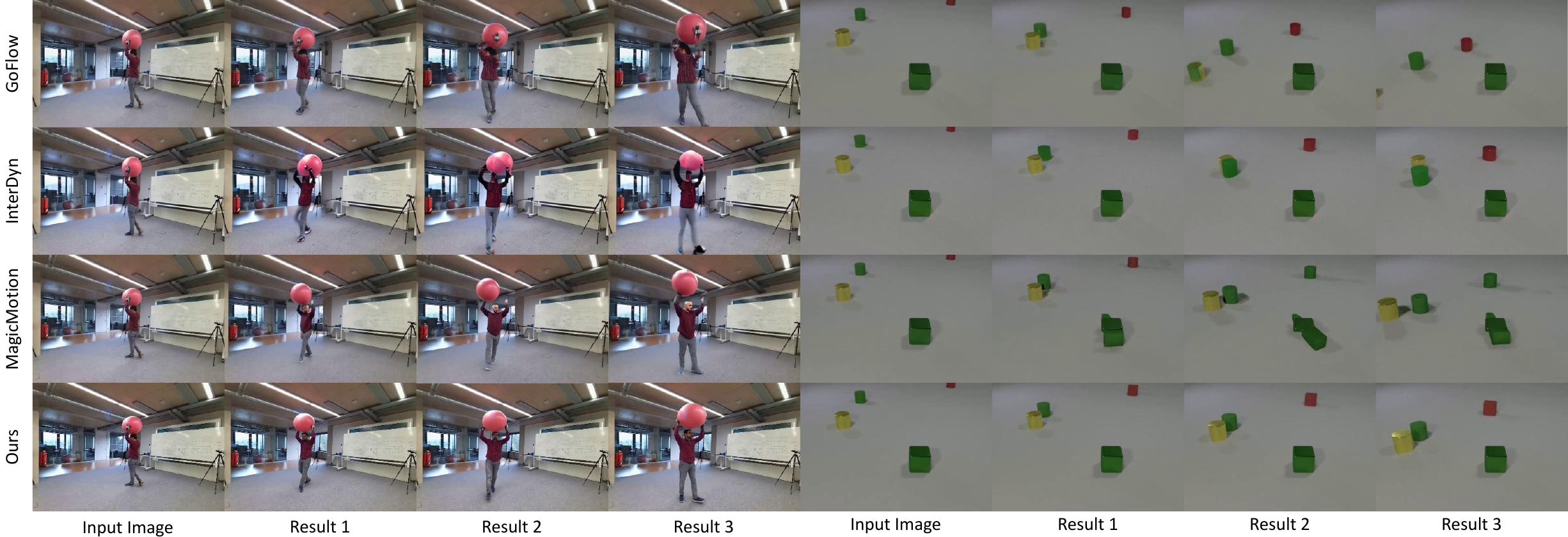}
    \caption{\textbf{Qualitative comparison on human-object interactions and object collision events.} Our method successfully captures the human-object interactions and object collision events.
    }
    \label{fig:hoi_comp}
\end{figure*}
We show that \OURS can successfully alter the motion of the controlled object by providing a mask.  
We observe force propagation not only between the controlled and uncontrolled objects, but also among the uncontrolled objects themselves. 
In Fig.~\ref{fig:results_clevrer} (Force Propagation), we show this propagation. In the single-object example (bottom), by controlling only the purple cube entering from the top, \OURS causes it to collide with and move the brown cube (controlled-uncontrolled interaction). Then, the brown cube collides with the gray cylinder and sets it into motion (uncontrolled-uncontrolled interaction).

We further show that modifying the control signal results in different outcomes, as shown in Fig.~\ref{fig:results_clevrer} (Control Signal Editing). 
In the first example (top), when two objects (the yellow cube and the purple cylinder) are controlled, the gray sphere first collides with the yellow cube and moves towards the purple cylinder, colliding with it. 
In the second example (bottom), we changed the control signal to only include the yellow cube, and the interactions of the gray sphere is also modified accordingly by our model.

In Table~\ref{tab:hoi_table} (CLEVRER), we compare our method with Go-with-the-Flow (GoFlow)~\cite{goflow}, InterDyn*~\cite{interdyn}, and MagicMotion~\cite{magicmotion}. Overall, the results show that the proposed intervention improves video-level realism while maintaining strong similarity to the ground truth. We consistently observe improvements in temporal realism (lower FVD), which aligns with our goal of generating temporally coherent dynamics.

Qualitative comparisons in Fig.~\ref{fig:hoi_comp} further support this conclusion. Baseline methods may produce plausible motion but often blur the causal structure of the event. In contrast, our method better preserves collision-driven changes and remain temporally stable after the interaction.
Refer to our video in supplemental material to observe the videos qualitatively.

\subsection{Human-Object Interaction with Human Mask Conditioning}
\label{sec:behave}

Besides altering the motion of rigid synthetic objects, \OURS can also guide non-rigid subjects with complex motion: humans. Although we provide only the mask of the moving subject, \OURS can generate the resulting human-object interactions, including grasping, object displacement aligned with hand or foot motion, and object position changes consistent with body rotation.
 
In Fig.~\ref{fig:results_behave}, we show interactions with three different objects. \OURS produces plausible human-object interactions by conditioning only on the subject mask, without requiring either object masks or explicit object motion guidance.

In Table~\ref{tab:hoi_table} (BEHAVE), we compare our method with the baselines and observe consistent improvements in video fidelity, while maintaining strong performance across reconstruction metrics. Fig.~\ref{fig:hoi_comp} provides a qualitative comparison. The baselines often produce noticeable artifacts under complex motions, whereas our method generates more temporally consistent results.
Refer to our video in supplemental material to observe the videos qualitatively.

\subsection{Discussion on Out-of-distribution Robustness and Efficiency}
\label{sec:ood_results}

\begin{figure*}[t]
  \centering
  \begin{minipage}[t]{0.48\textwidth}
    \centering
    \includegraphics[width=\textwidth]{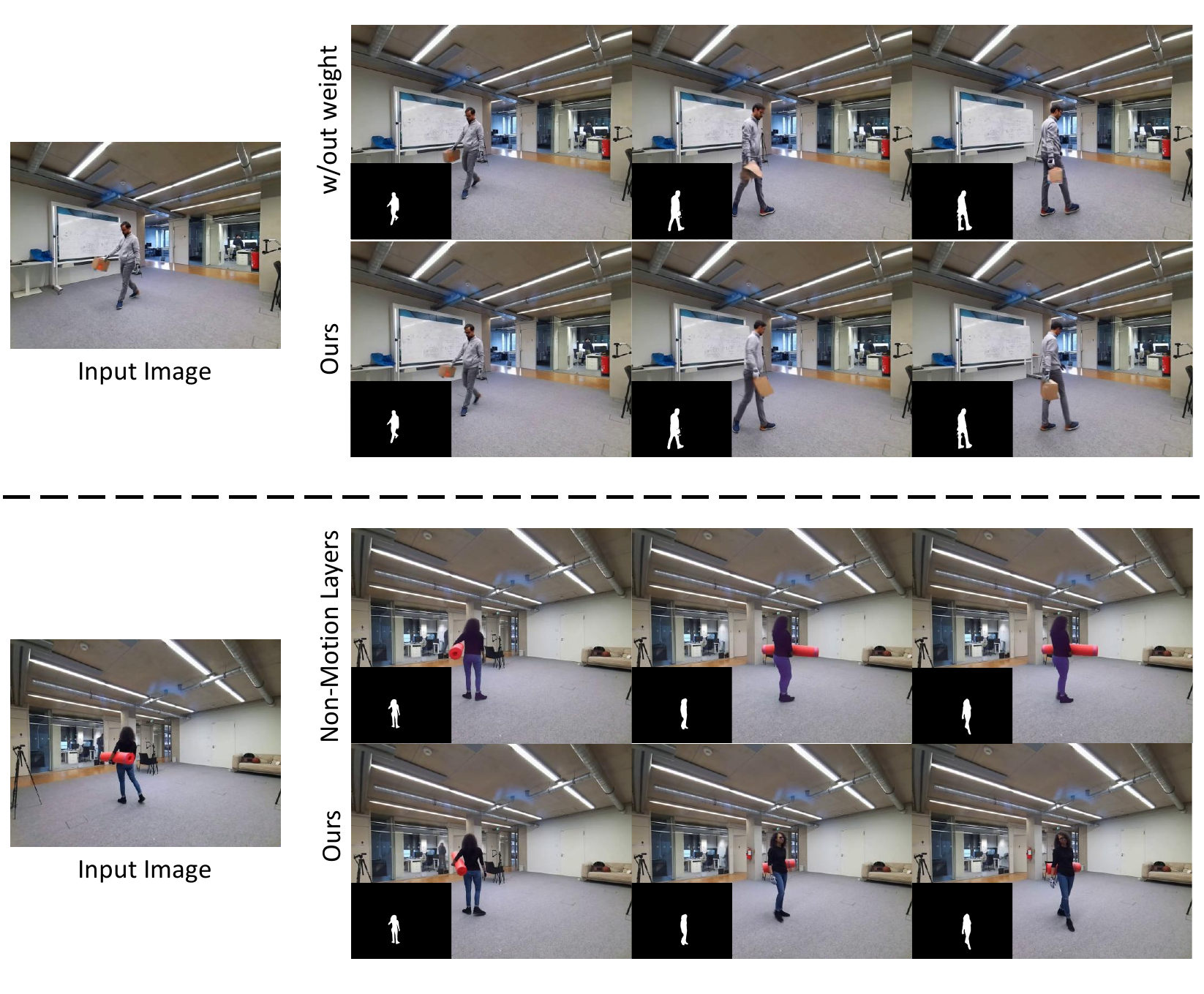}
    \caption{\textbf{Qualitative ablation.} Top: removing cosine-weighted latent injection introduces textural artifacts. Bottom: training LoRA on random non-motion layers weakens motion control.}
    \label{fig:ablation}
  \end{minipage}\hfill
  \begin{minipage}[t]{0.49\textwidth}
    \centering
    \includegraphics[width=\textwidth]{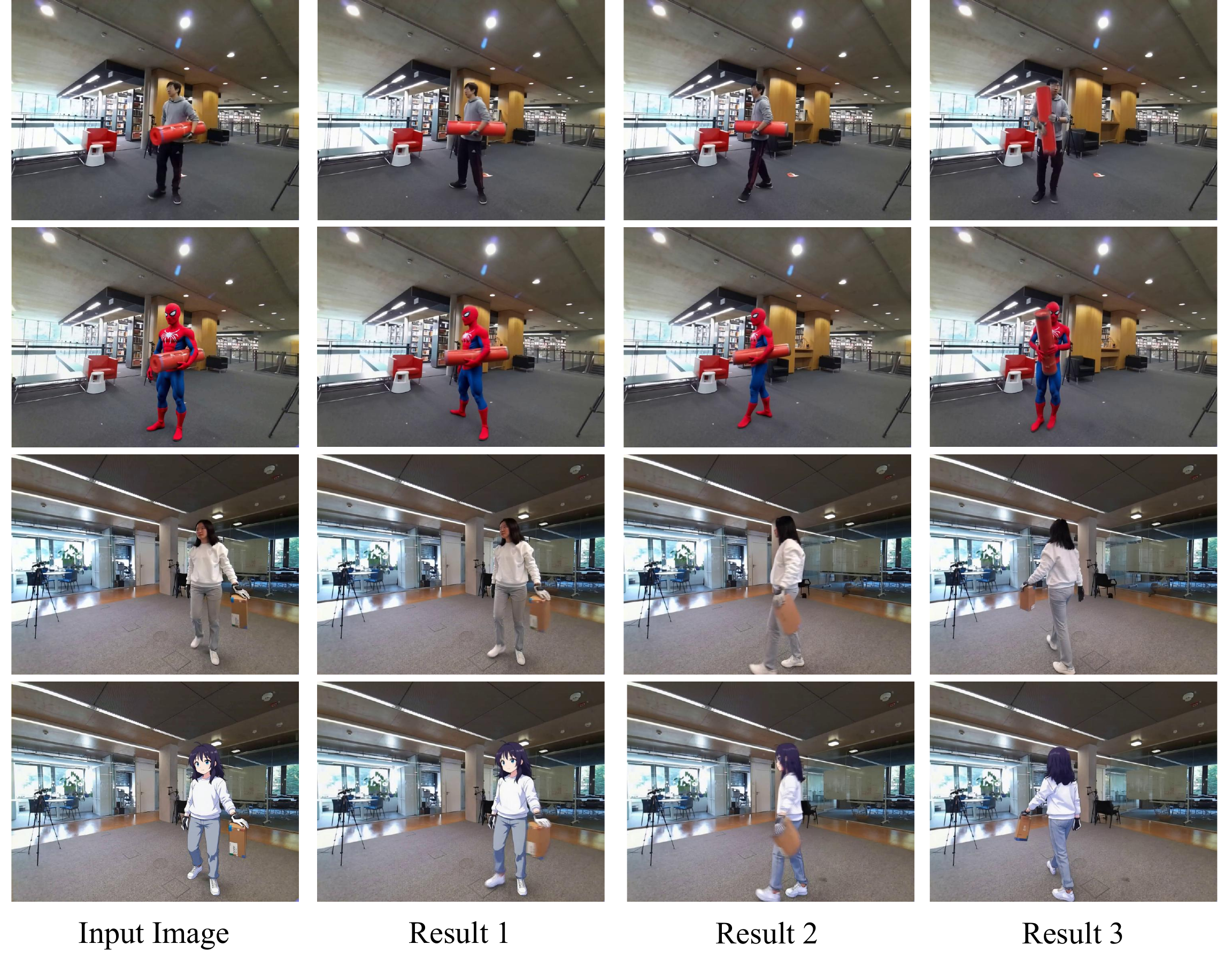}
    \caption{\textbf{Out-of-distribution edits.} We edit input images with a text-guided editor~\cite{flux1kontext} and show that generated interactions remain coherent on fictional characters.}
    \label{fig:edit}
  \end{minipage}
  \vspace{2mm}
\end{figure*}

We show that by applying text-guided edits to the input images, we can generate interactions for fictional or out-of-distribution characters, where Fig.~\ref{fig:edit} shows the out-of-distribution inputs remain compatible with our framework and the resulting videos preserve coherent motion. At the same time, we avoid the heavy conditioning used by recent mask-conditioned video generation baselines, which typically rely on a ControlNet-style branch that duplicates the backbone (e.g., a trainable copy of the frozen model), which adds 100\% extra parameters at both inference and training (see Tab.~\ref{tab:efficiency}). In contrast, our MaskAdapter is just two lightweight convolutional layers followed by a linear projection, adding only 1.84\% parameters at inference, while training uses compact LoRA updates (an additional 1.33\%), for a total of 3.17\% extra parameters during training. Overall, \OURS maintains qualitative robustness under appearance shifts while being markedly more parameter-efficient than prior mask-conditioned approaches.

\begin{table}[t]
    \centering
    \scalebox{0.99}{
    \begin{tabular}{@{}l|cccc|cccc@{}}
\toprule
& \multicolumn{4}{c}{\textbf{CLEVRER}} & \multicolumn{4}{c}{\textbf{BEHAVE}} \\
\cmidrule(lr){2-5}\cmidrule(lr){6-9}
Method & SSIM $\uparrow$ & PSNR $\uparrow$ & LPIPS $\downarrow$ & FVD $\downarrow$ & SSIM $\uparrow$ & PSNR $\uparrow$ & LPIPS $\downarrow$ & FVD $\downarrow$\\
\midrule
\mbox{GoFlow}~\cite{goflow} & 0.9117 & 26.56 & \textbf{0.1085} & 341.01 & 0.5044 & 19.21 & 0.1823 & 617.24 \\
\mbox{InterDyn*}~\cite{interdyn} & 0.8275 & 25.16 & 0.1632 & 358.02 & 0.5006 & 16.98 & 0.1893 & 720.14 \\
\mbox{MagicMotion}~\cite{magicmotion} & 0.7612 & 23.16 & 0.3315 & 466.51 & 0.7718 & \textbf{23.28} & 0.1191 & 694.41 \\
\midrule
Ours & \textbf{0.9252} & \textbf{27.12} & 0.1732 & \textbf{258.52} & \textbf{0.7940} & 22.99 & \textbf{0.0721} & \textbf{327.63} \\
\bottomrule
\end{tabular}
    }
    \vspace{2mm}
    \caption{
        \textbf{Quantitative comparison.} We compare our method with state-of-the-art controllable video generation models.
    }
    \label{tab:hoi_table}
\end{table}

\begin{table*}[t]
\vspace{-3mm}
  \centering
  \begin{minipage}[t]{0.43\textwidth}
    \centering
    \scalebox{0.75}{
    \begin{tabular}{lcc}
\toprule
Method & Inference & Training  \\
\midrule
InterDyn~\cite{interdyn} & 100 & 100 \\
MagicMotion~\cite{magicmotion} & 100 & 100 \\
Ours & 1.84 & 3.17 \\
\bottomrule
\end{tabular}
    }
    \vspace{2mm}
    \captionof{table}{\textbf{Efficiency comparison.} We report the percentage of additional parameters.}
    \label{tab:efficiency}
  \end{minipage} \hfill
  \begin{minipage}[t]{0.55\textwidth}
    \centering
    \scalebox{0.75}{
    \begin{tabular}{@{}lcccc@{}}
\toprule
Setting &  SSIM $\uparrow$ & PSNR$\uparrow$ & LPIPS $\downarrow$ & FVD $\downarrow$ \\
\midrule
LoRA on Non-\layer & 0.7787 & 21.81 & 0.0922 & 371.96 \\

w/o Cosine-weighted injection & 0.7938 & 22.99 & 0.0821 & 336.63 \\

Ours & \textbf{0.7940} & \textbf{22.99} & \textbf{0.0721} & \textbf{327.63} \\
\bottomrule
\end{tabular}
    }
    \vspace{2mm}
    \captionof{table}{
        \textbf{Quantitative Ablation Study.} We evaluate selective layer injection and cosine-weighted injection in the human-object interaction setting.
    }
    \label{tab:abl_table}
  \end{minipage}
  \vspace{-3mm}
\end{table*}

\subsection{Ablation Study}
\label{sec:ablation}
We ablate the effects of cosine-weighted latent injection, and restricting LoRA updates to Non-\layer. Fig.~\ref{fig:ablation} and Table~\ref{tab:abl_table} summarize the results.
A key design choice is adapting only \layer. To verify the layer selection, we swap \layer with an equal number of randomly chosen Non-\layer. As shown in Fig.~\ref{fig:ablation} (bottom-left), training on random Non-\layer loses precise controllability, consistent with the quantitative drop in Table~\ref{tab:abl_table}. We also ablate the cosine-weighted latent schedule by applying the residual $\Delta Z$ with a constant weight across all steps. As shown in Fig.~\ref{fig:ablation} (top row), this introduces undesirable textural artifacts and degrades all metrics (Table~\ref{tab:abl_table}), suggesting that overly strong late-step conditioning disrupts refinement of fine details. The cosine schedule mitigates this by emphasizing mask alignment early while preserving refinement capabilities in later denoising steps.
\section{Conclusion}
\label{sec:conclusion}

We presented \OURS, a controllable video generation framework that produces realistic interactive dynamics from a single binary steering mask sequence conditioned on an input image. We show that motion formation in Multi Modal Diffusion Transformers is concentrated in \layer, enabling targeted control. 
Our lightweight MaskAdapter injects mask derived latent residuals through \layer with cosine scheduling and layer-restricted LoRA, resulting precise motion and coherent interactions with minimal overhead.

A remaining challenge arises under extreme and prolonged occlusion, where objects disappear entirely and later re-emerge, making identity recovery difficult for the video model. \OURS demonstrates that precise control does not require complex inputs: temporally aligned mask guidance, applied at the right transformer layers, is enough to produce realistic scene dynamics. This formulation brings controllable video generation closer to applications where mask guidance can drive complex motion and interactions.
\bibliographystyle{splncs04}
\bibliography{main}
\clearpage
\appendix
\section{Implementation Details}
\label{sec:implementation}
We train models on two different datasets: one on CLEVRER~\cite{clevrer} and one on BEHAVE~\cite{behave}. This mirrors the experimental goal of isolating object-centric collision dynamics and human-centric interaction dynamics. Both models follow the same method described in the main paper, the only difference is the dataset. We train all experiments at 360p resolution. We optimize with AdamW~\cite{adamw} and a learning rate of $5 \times 10^{-5}$. Training proceeds in two stages: the MaskAdapter is trained for 1 epoch with the backbone frozen, after which we enable LoRA on the Motion Layers and fine-tune LoRA and the MaskAdapter jointly for 2 additional epochs. We use a batch size of two and train on a single NVIDIA H100 GPU.

\section{Base Model Details}
\OURS builds upon HunyuanVideo-I2V~\cite{hunyuanvideo}, a rectified-flow based I2V generation framework that uses an MMDiT architecture~\cite{flow-ode}.
Given an input image $I \in \mathbb{R}^{H \times W \times 3}$ and a text prompt, the image is first encoded into a latent representation using a pretrained 3DVAE encoder.
The generation process operates in the spatiotemporal latent space, initializing a noised video latent
$Z_t \in \mathbb{R}^{C \times \frac{T}{4} \times \frac{H}{8} \times \frac{W}{8}}$
where the temporal length follows the backbone constraint $T = 4n + 1$ (up to 129 frames).
In the I2V setting, $Z_{\text{img}}$ is injected into the first frame of the latent sequence to preserve identity and scene layout, and is also provided as an additional conditioning signal throughout generation, complementing the text condition.
\par
We select HunyuanVideo-I2V as our base model, because the MMDiT design results in rich attention maps across layers and modalities, providing a natural handle for diagnosing where subject-centric motion is formed and where controllability can be injected.
This motivates our method design, where we keep the base model mostly frozen and focus adaptation and conditioning on a small subset of \layer identified by our layer analysis.

\begin{figure*}[htbp]
  \centering
  \includegraphics[width=.9\textwidth]{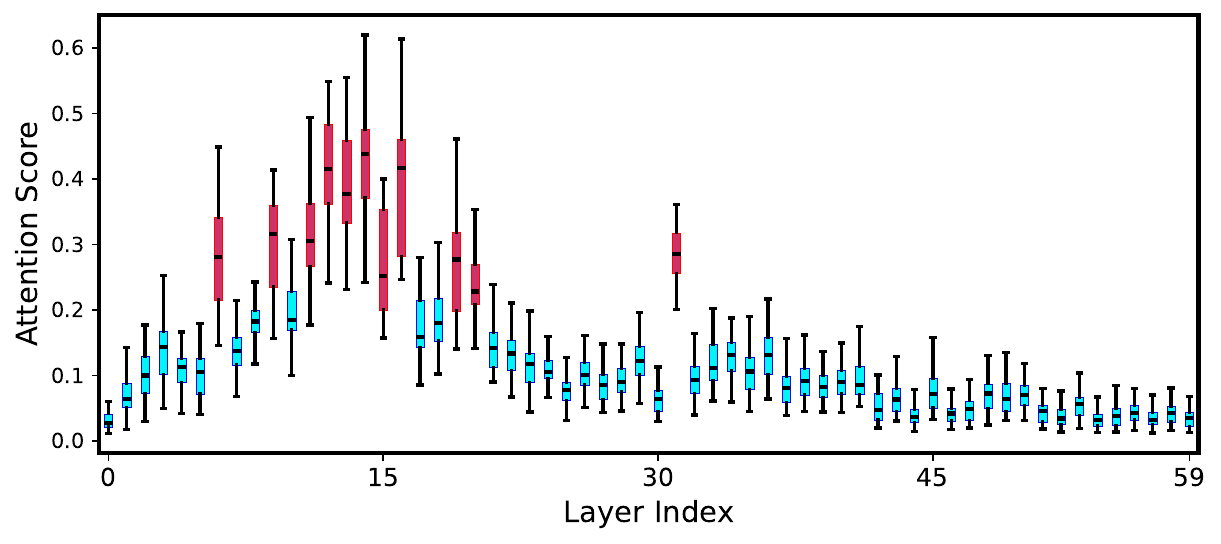}
  \caption{
  \textbf{Attention Score} visualization across the layers with error bars.}
  \label{fig:layer_score_box}
\end{figure*}

\section{Layer Analysis Details}
We conduct layer analysis using 50 samples from the BEHAVE~\cite{behave} dataset. We give simple prompts like "A man grabbing a backpack from the ground" or "A woman carrying a table", and calculate the mean of Text-to-Video and Video-to-Text attention with respect to the subject token (i.e man, woman). The detailed attention scores with
error bars are reported in Fig.~\ref{fig:layer_score_box}. It shows that attention scores are consistent across different samples where \layer consistently results in higher Attention Scores.

For the layer-skipping evaluations based on the selected \layer, we evaluate using 60 samples from the in-the-wild InterPose~\cite{interpose} dataset. 
We calculate $\mathcal{J}$, $\mathcal{F}$~\cite{Perazzi2016}, and HOTA~\cite{luiten2020trackeval,luiten2020IJCV} scores. 
Specifically, $\mathcal{J}$ is the Jaccard index (IoU) between the pseudo ground-truth mask and predicted mask which represents region similarity. $\mathcal{F}$ is the contour accuracy, computed from contour precision/recall, emphasizing shape alignment. 
HOTA (Higher Order Tracking Accuracy) is a standard multi object tracking (MOT) metric that aids in measuring identity consistency, we report HOTA as an extra validation to our approach.

\section{Mask Robustness Evaluations}

We also evaluate how robust our model is to variations in mask inputs.

\noindent
\textbf{Mask Frame Skipping.} To examine the sensitivity of \OURS to incomplete mask guidance, we perform a mask robustness study in which we progressively remove portions of the input mask sequence at inference time. Specifically, starting from the original binary mask sequence, we evaluate four settings: using all masks, and retaining only one mask every 4, or 8 frames. For the skipped frames, the corresponding masks are blacked out, so that the model receives no subject guidance at those timesteps. 
Because retaining only one mask every eight frames can leave the compressed latent space without any control signal, we propagate the subsequent mask to the preceding skipped frame.  
We conduct this evaluation on a subset of 10 samples from the evaluation set.
\textit{Please refer to the video results in the project webpage}.
When every mask is provided, the generated motion follows the intended trajectory most accurately. As masks are skipped more aggressively, we observe mild drift and reduced spatial precision in some frames, but the overall dynamics remain coherent in most cases. Even in the most sparse setting, where only one mask is retained every 8 frames, the model continues to recover temporally plausible motion rather than collapsing completely. 
Overall, these results indicate that \OURS is reasonably robust to temporally sparse mask conditioning, while still benefiting from denser mask supervision when precise control is required.
\\

\noindent\textbf{In the Wild Setting.} To further evaluate generalization beyond the benchmark datasets, we consider an in-the-wild setting using online videos obtained from the InterPose dataset~\cite{interpose}. These videos exhibit larger appearance variation and less constrained scene structure than the training data (i.e., BEHAVE~\cite{behave}), which tests whether the same mask-to-motion principle remains effective under more diverse real-world conditions. 
\textit{Please refer to the qualitative results in the project webpage}.

\begin{figure*}[t]
  \centering
  \includegraphics[width=.9\textwidth]{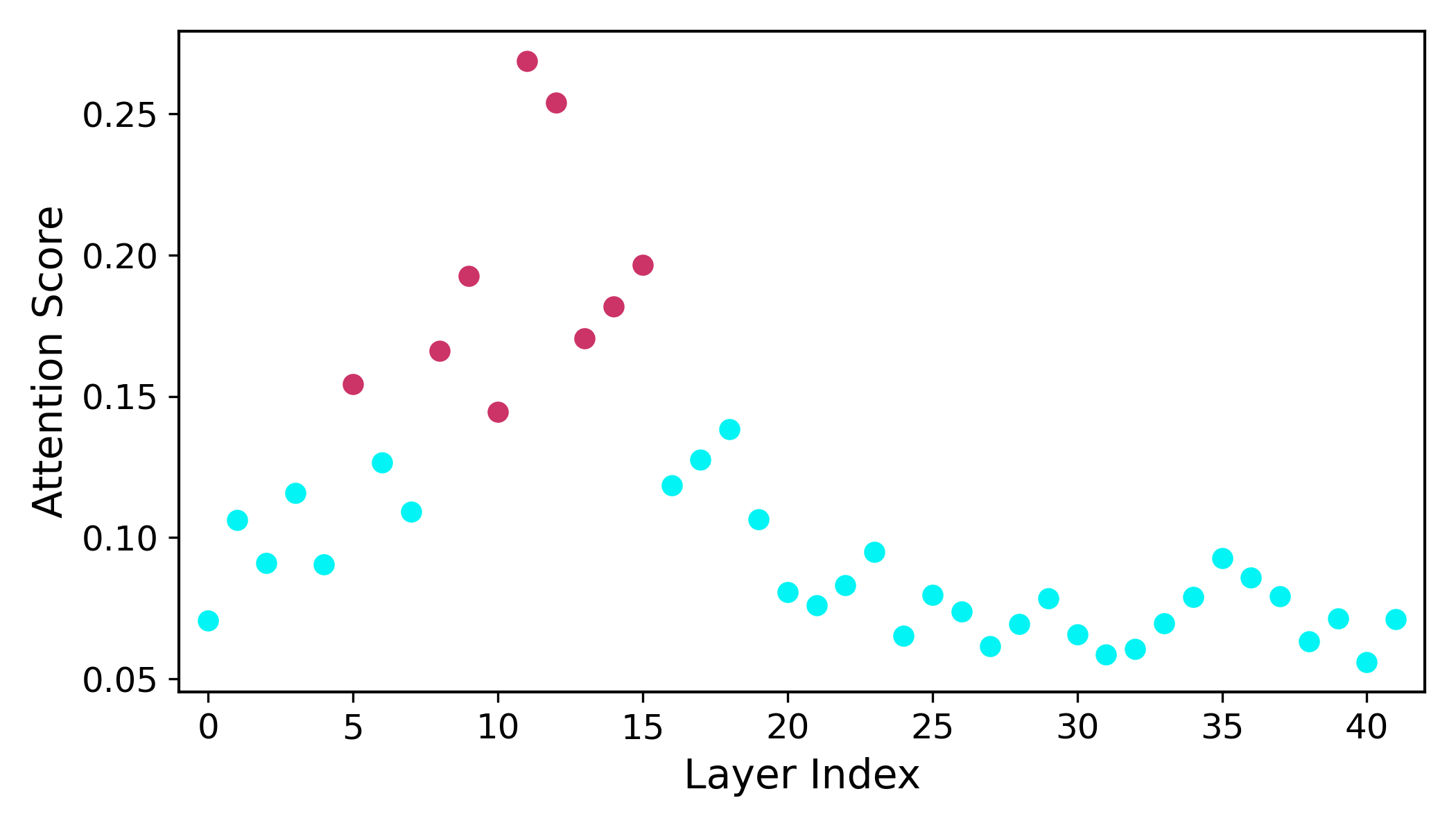}
  \caption{
  \textbf{Attention Score} visualization of CogVideoX~\cite{yang2024cogvideox}.}
  \label{fig:layer_score_cogvid}
\end{figure*}

\begin{table}[t]
\centering
    \scalebox{0.99}{
    \begin{tabular}{l cc cc c c}
\hline
Method & $\mathcal{J}\uparrow$& $\mathcal{F}\uparrow$ & $\mathcal{J}\&\mathcal{F}\uparrow$ & HOTA$\uparrow$\\
\midrule
Skip Motion Layers     & 39.4 & 32.0 & 35.7 & 36.5\\
Skip Non-Motion Layers & \textbf{53.5} & \textbf{48.3} & \textbf{50.9} & \textbf{55.0}\\
\bottomrule
\end{tabular}
    }
    \caption{
        \textbf{Layer analysis for CogVideoX~\cite{yang2024cogvideox}.} We generate videos skipping the randomly selected 3 \layer{} and Non-\layer{}, and report the tracking metrics for CogvideoX DiT model.
    }
\label{tab:JF_score_cogvid}
\end{table}

\section{Layer Analysis for CogVideoX}

In the paper, our layer analysis focused on HunyuanVideo \cite{hunyuanvideo}, which is a recent DiT model.
Here, we repeat the layer analysis for another DiT model CogVideoX-5B-I2V~\cite{yang2024cogvideox} following the same procedure as in the main paper. Specifically, we compute the Attention Score for each transformer block by measuring the alignment between the subject mask and the corresponding subject-token attention maps across inference steps, and then rank the layers accordingly. Based on this ranking, we define Motion Layers and Non-Motion Layers in CogVideoX (see Fig.~\ref{fig:layer_score_cogvid}), and perform the same controlled layer-skipping experiment used for the MMDiT backbone. For each generated sample, we keep the initial random noise fixed and compare the original generation with two additional variants obtained by skipping randomly selected Motion Layers or randomly selected Non-Motion Layers. This allows us to isolate whether the layers identified by the attention-based ranking are in fact more critical to motion formation in CogVideoX as well. We measure tracking metrics and report the results in Tab.~\ref{tab:JF_score_cogvid}, which shows that skipping Motion Layers causes significantly more disruption to video quality.

\section{Applications}
In this section, we highlight several potential applications of our model.
\textit{Please refer to the project webpage for qualitative examples of these applications.}
All results are produced using the checkpoint trained on BEHAVE~\cite{behave}, demonstrating the model's ability to generalize beyond its training setting.

\subsection{Text-Guided Human Animation}
In combination with existing open-source models, our approach can be used for text-guided human animation.
Given a reference human image, we first estimate a human mesh using HMR2.0~\cite{4dhuman}. We then employ the motion editing and inpainting capabilities of Motion Diffusion Model (MDM)~\cite{mdm} to generate a motion sequence from a text prompt. Finally, we render the animated mesh and extract the corresponding binary mask sequence, which is used as control input for our model.

\subsection{Motion-to-Video Animation}
Our method can also be used to convert generated motion into realistic videos.
We first use MDM~\cite{mdm} to synthesize a motion sequence, then render the animated human mesh to obtain a binary mask sequence. Next, we transform the first frame with an image-to-image translation model, Gemini Nano Banana~\cite{gemini3}, and feed both the transformed first frame and the mask sequence into \OURS to generate the final realistic video.

\subsection{Object Animation and Manipulation}
Beyond human animation, our model can also be applied to object manipulation tasks.
For instance, given an input image of an object, users can provide a sequence of handcrafted masks representing the desired motion or transformation. These masks may be handcrafted manually, or obtained simply through specifying the translation and rotation. Then, \OURS can be used to generate a video that follows the specified object movement.

\subsection{Style Transfer}
To demonstrate that \OURS can generalize to out-of-distribution subjects and visual domains, we first apply image-to-image translation to the input image~\cite{gemini3,flux1kontext}. We then animate the translated image using \OURS, conditioned on the stylized first frame and its corresponding temporal binary mask sequence.
\end{document}